\documentclass[11pt,letterpaper]{article}
\usepackage{authblk}
\usepackage{emnlp2017}
\usepackage{times}
\usepackage{latexsym}
\usepackage{amssymb}
\usepackage{multirow}
\usepackage{graphicx}
\usepackage{url}
\usepackage{amsmath}
\emnlpfinalcopy

\def\emnlppaperid{30}

\title{An Error-Oriented Approach to Word Embedding Pre-Training}

\author[1]{Youmna Farag}
\author[1,2]{Marek Rei}
\author[1,2]{Ted Briscoe}
\affil[1]{Computer Laboratory, University of Cambridge, United Kingdom}
\affil[2]{The ALTA Institute, Cambridge, United Kingdom}
\affil[ ]{\tt {\small\{youmna.farag,marek.rei,ted.briscoe\}@cl.cam.ac.uk}}

\date{}

\begin{document}
\maketitle
\begin{abstract}
We propose a novel word embedding pre-training approach that exploits writing errors in learners' scripts. We compare our method to previous models that tune the embeddings based on script scores and the discrimination between correct and corrupt word contexts in addition to the generic commonly-used embeddings pre-trained on large corpora. The comparison is achieved by using the aforementioned models to bootstrap a neural network that learns to predict a holistic score for scripts. Furthermore, we investigate augmenting our model with error corrections and monitor the impact on performance. Our results show that our error-oriented approach outperforms other comparable ones which is further demonstrated when training on more data. Additionally, extending the model with corrections provides further performance gains when data sparsity is an issue. 
\end{abstract}

\section{Introduction} 
Assessing students' writing plays an inherent pedagogical role in the overall evaluation of learning outcomes. Traditionally, human graders are required to mark essays, which is cost- and time-inefficient, especially with the growing numbers of students. Moreover, the evaluation process is subjective, which leads to possible variations in the awarded scores when more than one human assessor is employed. To remedy this, the automated assessment (AA) of writing has been motivated in order to automatically evaluate writing competence and hence not only reduce grader workload, but also bypass grader inconsistencies as only one system would be responsible for the assessment. Numerous AA systems have been developed for research purposes or deployed for commercial use, including Project Essay Grade (PEG) ~\cite{page2003project}, e-Rater ~\cite{attali2006automated}, Intelligent Essay Assessor (IEA)~\cite{landauer2003automated} and Bayesian Essay Test Scoring sYstem (BETSY) ~\cite{rudner2002automated} among others. They employ statistical approaches that exploit a wide range of textual features. 

A recent direction of research has focused on applying deep learning to the AA task in order to circumvent the heavy feature engineering involved in traditional systems. Several neural architectures have been employed including variants of Long Short-Term Memory (LSTM)~\cite{alikaniotis2016automatic,taghipour2016neural} and Convolutional Neural Networks (CNN)~\cite{dong-zhang-2016}. They were all applied to the Automated Student Assessment Prize (ASAP) dataset, released in a Kaggle contest\footnote{https://www.kaggle.com/c/asap-aes/}, which contains essays written by middle-school English speaking students. On this dataset, neural models that only operate on word embeddings outperformed state-of-the-art statistical methods that rely on rich linguistic features~\cite{yannakoudakis2011new,phandi2015flexible}.

The results obtained by neural networks on the ASAP dataset demonstrate their ability to capture properties of writing quality without recourse to handcrafted features. However, other AA datasets pose a challenge to neural models and they still fail to beat state-of-the-art methods when evaluated on these sets. An example of such datasets is the First Certificate in English (FCE) set where applying a rank preference Support Vector Machine (SVM) trained on various lexical and grammatical features achieved the best results~\cite{yannakoudakis2011new}. This motivates further investigation into neural networks to determine what minimum useful information they can utilize to enhance their predictive power.

Initializing neural models with contextually rich word embeddings pre-trained on large corpora ~\cite{mikolov2013distributed,pennington2014glove,turian2010word} has been used to feed the networks with meaningful embeddings rather than random initialization. Those embeddings are generic and widely employed in Natural Language Processing (NLP) tasks, yet few attempts have been made to learn more task-specific embeddings. For instance, \citet{alikaniotis2016automatic} developed \textit{score-specific word embeddings} (SSWE) to address the AA task on the ASAP dataset. Their embeddings are constructed by ranking correct ngrams against their ``noisy" counterparts, in addition to capturing words' informativeness measured by their contribution to the overall score of the essay. 

We propose a task-specific approach to pre-train word embeddings, utilized by neural AA models, in an error-oriented fashion. Writing errors are strong indicators of the quality of writing competence and good predictors for the overall script score, especially in scripts written by language learners, which is the case for the FCE dataset. For example, the Spearman's rank correlation coefficient between the FCE script scores and the ratio of errors is $-0.63$ which is indicative of the importance of errors in writing evaluation:
\begin{equation*}
 \text{ratio of errors} = \frac{\text{number of erroneous script words}}{\text{script length}}
\end{equation*}
This correlation could even be higher if error severity is accounted for as some errors could be more serious than others. Therefore, it seems plausible to exploit writing errors and integrate them into AA systems, as was successfully done by \citet{yannakoudakis2011new} and \citet{rei2016compositional}, but not by capturing this information directly in word embeddings in a neural AA model.

Our pre-training model learns to predict a score for each ngram based on the errors it contains and modifies the word vectors accordingly. The idea is to arrange the embedding space in a way that discriminates between ``good" and ``bad" ngrams based on their contribution to writing errors. Bootstrapping the assessment neural model with those learned embeddings could help detect wrong patterns in writing which should improve its accuracy of predicting the script's holistic score.

We implement a CNN as the AA model and compare its performance when initialized with our embeddings, tuned based on natural writing errors, to the one obtained when bootstrapped with the SSWE, proposed by \citet{alikaniotis2016automatic}, that relies on random noisy contexts and script scores. Furthermore, we implement another version of our model that augments ngram errors with their corrections and investigate the effect on performance. Additionally, we compare the aforementioned pre-training approaches to the commonly used embeddings trained on large corpora (Google or Wikipedea). The results show that our approach outperforms other initialization methods and augmenting the model with error corrections helps alleviate the effects of data sparsity. Finally, we further analyse the pre-trained representations and demonstrate that our embeddings are better at detecting errors which is inherent for AA.

\begin{figure*}[h]
\centering
\includegraphics[scale=0.5]{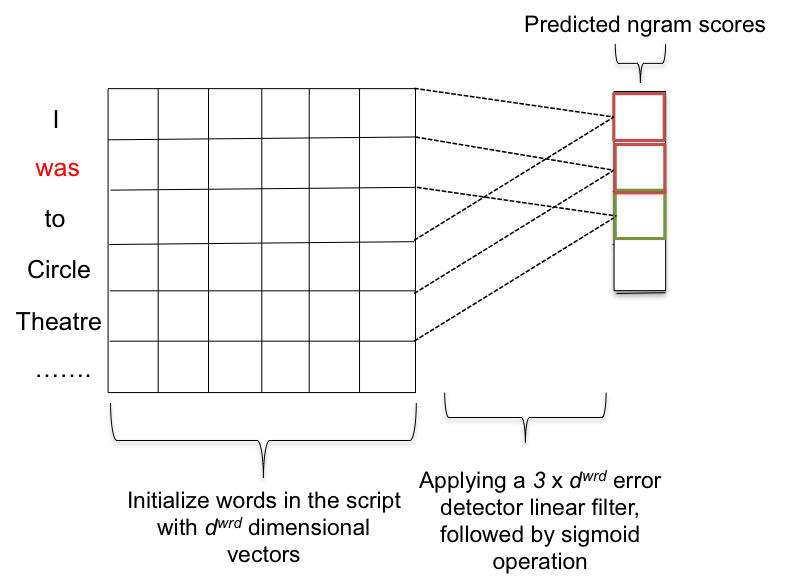}
\caption{Error-specific Word Embeddings (ESWE).}
\label{figure1}
\end{figure*}

\section{Related Work}
There have been various attempts to employ neural networks to assess the essays in the ASAP dataset.
\citet{taghipour2016neural} compared the performance of a few neural network variants and obtained the best results with an LSTM followed by a \textit{mean over time} layer that averages the output of the LSTM layer. \citet{alikaniotis2016automatic} assessed the same dataset by building a bidirectional double-layer LSTM which outperformed Distributed Memory Model of Paragraph Vectors
(PV-DM)~\cite{le2014distributed} and Support Vector Machines (SVM) baselines. \citet{dong-zhang-2016} implemented a CNN where the first layer convolves a filter of weights over the words in each sentence followed by an aggregative pooling function to construct sentence representations. Subsequently, a second filter is applied over sentence representations followed by a pooling operation then a fully-connected layer to predict the final score. Their CNN was applied to the ASAP dataset and its efficacy in in-domain and domain-adaptation essay evaluation was demonstrated in comparison to traditional state-of-the-art baselines.

Several AA approaches in the literature have exploited the ``quality" or ``correctness" of ngrams as a feature to discriminate between good and poor essays.
\citet{phandi2015flexible} defined good essays as the ones with grades above or equal to the average score and the rest as poor ones. They calculated the Fisher scores~\cite{fisher1922interpretation} of ngrams and selected $201$ with the highest scores as \textit{``useful ngrams"}. Similarly, they generated correct POS ngrams from grammatically correct texts, classified the rest as ``bad POS ngrams" and used them along with the useful ngrams and other shallow lexical features as bag-of-words features. They applied Bayesian linear ridge regression (BLRR) and SVM regression for domain-adaptation essay scoring using the ASAP dataset. 
\citet{alikaniotis2016automatic} applied a similar idea; in their SSWE model, they trained word embeddings to distinguish between correct and noisy contexts in addition to focusing more on each word's contribution to the overall text score. Bootsrapping their LSTM model with those embeddings offered further performance gains. 

Other models have directly leveraged error information exhibited in text. For example, \citet{yannakoudakis2011new} demonstrated that adding an \textit{``error-rate"} feature to their SVM ranking model that uses a wide range of lexical and grammatical writing competence features further improves the AA performance. They calculated the error-rate using the error annotations in the Cambridge Learner Corpus (CLC) in addition to classifying a trigram as erroneous if it does not occur in the large ukWaC corpus~\cite{ferraresi2008introducing} or highly scoring CLC scripts.
 \citet{rei2016compositional} proposed a bidirectional LSTM for error detection in learner data, where the model predicts the probability of a word being correct for each word in text. As an extension to their experiment, they incorporated the average predicted probability of word correctness as an additional feature to the self-assessment and tutoring system (SAT)~\cite{andersen2013developing} that applied a supervised ranking perceptron to rich linguistic features. Adding their correctness probability feature successfully enhanced the predictive power of the SAT.

\begin{figure*}[h]
\centering
\includegraphics[scale=0.5]{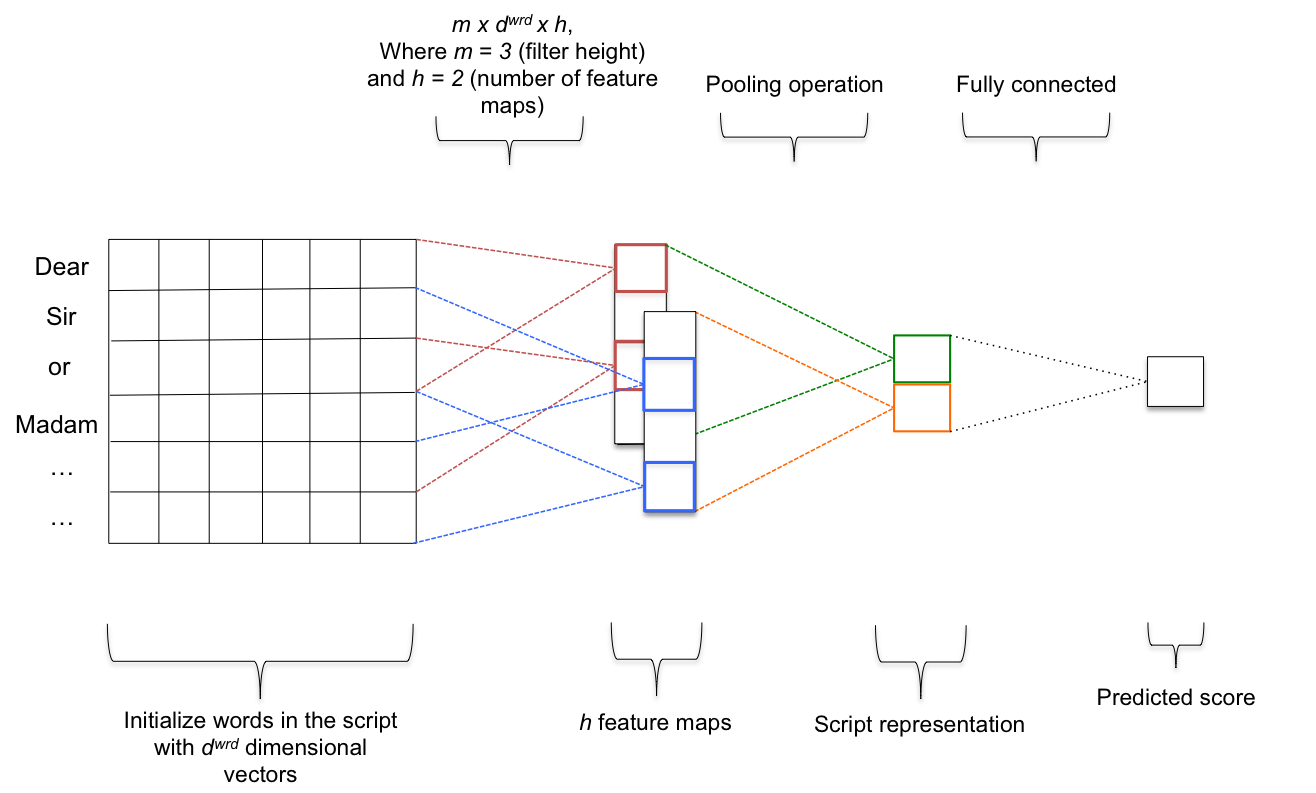}
\caption{A CNN for AA where the final score is predicted by applying a convolutional operation followed by a pooling function.}
\label{figure2}
\end{figure*}
\section{Approach}
\subsection{Word Embedding Pre-training}
\label{pretrain}
In this section, we describe three different neural networks to pre-train word representations: the model implemented by \citet{alikaniotis2016automatic} and the two error-oriented models we propose in this work. The models' output embeddings -- referred to as \textit{AA-specific} embeddings -- are used later to bootstrap the AA system.
\vspace{0.2cm} \\ 
\noindent{\bf Score-specific Word Embeddings (SSWE).}
We compare our pre-training models to the SSWE developed by \citet{alikaniotis2016automatic}. Their method is inspired by the work of~\citet{collobert2008unified} which learns word representations by distinguishing between a target word's context (window of surrounding words) and its noisy counterparts. These counterparts are generated by replacing the target word with a randomly selected word from the vocabulary. The network is trained to rank the positive correct contexts higher than the negative corrupt ones. 
Additionally, the model is augmented with score specific information to focus on the informative words that contribute to the overall score of essays rather than the frequent words that occur equally in good and bad essays. They optimize the overall loss function as a weighted sum of the ranking loss between correct and noisy ngrams and the score specific loss:
\begin{equation}
\label{ssweloss}
\begin{split}
Loss_{(SSWE)} &= \alpha \cdot loss_{(ranking)} \\
&+(1 - \alpha) \cdot loss_{(score)}
\end{split}
\end{equation}
where $\alpha$ is a hyperparameter. In their experiment, they set $\alpha$ to $0.1$ giving most of the weight to score-related information.  
\vspace{0.2cm} \\ 
\noindent{\bf Error-specific Word Embeddings (ESWE).}
We propose a model that fine-tunes the embedding space using a supervised method that leverages the errors appearing in the training data. It modifies the embedding space to discriminate between erroneous ngrams and correct ones. The core difference between this approach and SSWE is that it relies on the writing errors occurring naturally in the training data instead of randomly generating incorrect ngrams or capturing words' informativeness. The motivation for adopting this approach is twofold. First, we believe that the model could learn more useful AA features from actual errors rather than introducing random contexts that are unlikely to happen. Second, SSWE ignores the frequent words as they have less predictive power (they are used equally in highly and lowly scored texts). However, despite the fact that frequent words (e.g. function words) carry less topical information than content ones, the errors associated with them constitute a substantial portion of the errors committed by non-native English speakers. For instance, \textit{determiner errors} account for more than $9\%$ of the total errors in public FCE training data. Therefore, learning representations from both function and content word errors in their contexts could be advantageous. 

The ESWE model predicts error scores for word ngrams. First, we demonstrate how the true error scores for ngrams are calculated and second, we describe the approach applied to estimate these scores. Each word $w_i$ in a training document is given an error indicating score $e_i\in\{1,0\}$ based on whether it is part of an error or not, respectively. Subsequently, an ngram gold score ($n\_score$) is calculated based on the sum of the errors it contains as follows:
\begin{equation}
\label{scoreeq}
n\_score = \frac{1}{1+\sum^n_i{e_i}}
\end{equation}
where $n$ is the ngram length. For the model to estimate the ngram scores, a convolutional operation is applied as depicted in Figure~\ref{figure1}. First, each word is mapped to a unique vector $v_i^{wrd}\in\mathbb{R}^{d^{wrd}}$ retrieved from an embedding space $E\in\mathbb{R}^{|V|\times d^{wrd}}$, where $|V|$ is the vocabulary size. Consequently, an ngram is represented as a concatenation of its word vectors $v^{ng} = [v_i^{wrd};...;v_{i+n-1}^{wrd}]$. Scoring the ngrams is accomplished by sliding a convolutional linear filter $W^{e} \in \mathbb{R}^{n\times d^{wrd}}$ -- hereafter \textit{error filter}\footnote{We also refer to the window size used in SSWE as error filter for simplicity.} -- over all the ngrams in the script, followed by a sigmoid non-linearity to map the predicted score to a $[0,1]$ probability space:
\begin{equation}
\hat{n\_score} = \sigma(W^{e} \cdot v^{ng})
\end{equation}
where $\sigma$ is the sigmoid function.\footnote{Biases are removed from equations for simplicity.} The error filter should work as an error detector that evaluates the correctness of words given their contexts and arranges them in the embedding space accordingly. For optimization, the sum of squared errors loss is minimized between the gold ngram scores and the estimated ones and the error gradients are backpropagated to the embedding matrix $E$ building the ESWE space:
\begin{equation}
\label{loss}
Loss = \sum_k{(n\_score_k - \hat{n\_score_k})^2}
\end{equation}
where $k$ is the ngram index.
\vspace{0.2cm} \\ 
\noindent{\bf Error-correction-specific Word Embeddings (ECSWE).} 
As an extension to ESWE, we propose augmenting it with the errors' corrections as follows.
We build a corrected version of each script by replacing all its errors with their suggested corrections and train the ESWE model using the corrected scripts together with the original ones. In the corrected version, all the ngrams are given $e_i = 0$ and consequently, $n\_score = 1$ according to Equation~\ref{scoreeq}. All the above ESWE equations are applied and the loss for each script is calculated as the sum of both the loss of the original script and its corrected version (Equation~\ref{loss} applied to obtain both). The motivation for this model is twofold. First, it could enrich the embedding space by allowing the model to learn from faulty ngrams and their correct counterparts (both occur naturally in text) and construct ECSWE which is a modified version of ESWE that is more capable of distinguishing between good and bad contexts. Second, it could alleviate the effects of data sparsity, when training on small datasets, by learning from more representations.\footnote{We refer to ESWE and ECSWE as \textit{error-oriented} models.}
\begin{table}[]
\begin{tabular}{c|c|c|c}
\hline
Model                                                                                & Dataset & Error       & Script      \\ \hline
\multirow{2}{*}{\begin{tabular}[c]{@{}c@{}}Google Word2Vec \\ \& GloVe\end{tabular}} & FCE    & \multirow{2}{*}{-} & \multirow{2}{*}{3} \\ \cline{2-2}
                                                                                     & FCE$_{\textrm{ext}}$  &                    &                    \\ \hline
\multirow{2}{*}{\begin{tabular}[c]{@{}c@{}}SSWE, ESWE \&\\ ECSWE\end{tabular}}         & FCE  & 3                  & 3                  \\ \cline{2-4} 
                                                                                     & FCE$_{\textrm{ext}}$    & 9                  & 9                  \\ \hline
\end{tabular}
\centering
\caption{Error and script refer to their filter sizes. For each of the $5$ pre-training models on the two datasets, the error filter size is displayed (if applicable) along with the script filter size used in the AA network initialized with the embeddings on the left. FCE refers to the public FCE.}
\label{table1}
\end{table}

\subsection{AA Model}
\label{essayscoring}
The previous section discusses pre-training approaches for word embeddings that are later used to initialize the AA model.
For this model, we use a second CNN to predict a holistic score for the script (Figure~\ref{figure2}) as follows. Each word in an input script is initialized with its vector $v_i^{wrd'}\in\mathbb{R}^{d^{wrd}}$ from a pre-trained embedding matrix, resulting in a script embedding $[v_1^{wrd'};....;v_l^{wrd'}] $, where $l$ is the length of the script. A convolutional filter $W^s \in \mathbb{R}^{m \times d^{wrd} \times h}$ is slid over all the script's subsequences to generate the feature maps $M \in \mathbb{R}^{h\times (l - m + 1)} $, where $m$ is the filter height (window size) and $h$ is the number of the output feature maps. We refer to this filter as the \textit{script filter}.
Previously, for the error filter used in the ESWE and ECSWE approaches, $h$ was set to $1$ which represents the predicted ngram score ($\hat{n\_score}$), whereas here, the system extracts various contextual features from each ngram as a pre-step towards predicting the script's score, hence setting $h$ to a large value. The convolutional operation is followed by a \textit{ReLU} non-linearity to capture more complex linguistic phenomena:\footnote{Initial experimentation showed that \textit{ReLU} performs better than \textit{tanh} in the AA model.}
\begin{equation}
M_i = ReLU(W^s \cdot v_{i:i+m-1}^{wrd'})
\end{equation}
\begin{equation}
M = [M_1, M_2,...M_{l-m+1}]
\end{equation}
Subsequently, an average pooling function is applied to the output feature maps in order to select the useful features and unify the scripts' representations to a vector $S \in \mathbb{R}^{h}$ of fixed length.
Finally, the last layer of the network is a fully connected one by applying linear regression to the script representation in order to predict the final score: 
\begin{equation}
\hat{s\_score} = W^{reg} \cdot S
\end{equation}
where $W^{reg}\in \mathbb{R}^{h}$ is a learned parameter matrix. The network optimizes the sum of squared errors loss between the scripts' predicted scores and the gold ones.

\begin{table*}[]
\centering
\begin{tabular}{|c|c|c|c|}
\hline
Bootstrapping Model                & Pearson ($r$) & Spearman ($\rho$) & RMSE  \\ \hline
Google Word2Vec 300d & 0.488   & 0.446    & 5.339 \\ \hline
GloVe 50d            & 0.475   & 0.427    & 5.308 \\ \hline
SSWE                 & 0.494   & 0.445    & 5.182 \\ \hline
ESWE                  & 0.521   & 0.481    & 5.194 \\ \hline
ECSWE                 & \textbf{0.538}   & \textbf{0.499}    & \textbf{5.033} \\ \hline
\end{tabular}
\caption{AA results when bootstrapped from different word embeddings and trained on public FCE. The bold values indicate the best results.}
\label{table2}
\end{table*}
\section{Experimental Setup}
\noindent{\bf Baselines.} 
We compare our error-oriented approaches to the SSWE model as well as generic pre-trained models commonly used to initialize neural networks for different NLP tasks. The generic models are trained on large corpora to capture general semantic and syntactic regularities, hence creating richer, more meaningful word vectors, as opposed to random vectors. In particular, Google News Word2Vec ($d^{wrd} = 300$)~\cite{mikolov2013distributed} and GloVe ($d^{wrd} = 50$)~\cite{pennington2014glove} pre-trained models are used. Google Word2Vec\footnote{https://code.google.com/archive/p/word2vec/} is a Skip-gram model that learns to predict the context of a given word. It is trained on Google News articles which contain around $100$ billion words with $3$ million unique words. On the other hand, GloVe\footnote{https://nlp.stanford.edu/projects/glove/} vectors are learned by leveraging word-word cooccurrence statistics in a corpus. We use the GloVe embeddings trained on a 2014 Wikipedia dump in addition to Gigaword 5 with a total of $6$ billion words.
\vspace{0.2cm} \\ 
\noindent{\bf Evaluation.} 
We replicate the SSWE model, implement our ESWE and ECSWE models, use Google and GloVe embeddings and conduct a comparison between the $5$ initilization approaches by feeding their output embeddings to the AA system from Section~\ref{essayscoring}. All the models are implemented using the open-source Python library Theano~\cite{theano}. For evaluation, we calculate Spearman's rank correlation coefficient ($\rho$), Pearson's product-moment correlation coefficient ($r$) and root mean square error ($RMSE$) between the final predicted script scores and the ground-truth values~\cite{yannakoudakis2015evaluating}. 
\vspace{0.2cm} \\ 
\noindent{\bf Dataset.} 
For our experiments, we use the FCE dataset~\cite{yannakoudakis2011new} which consists of exam scripts written by English learners of upper-intermediate proficiency and graded with scores ranging from $1$ to $40$.\footnote{We only evaluate on FCE and not the ASAP dataset because the latter does not contain error annotations.}  
 Each script contains two answers corresponding to two different prompts asking the learner to write either an article, a letter, a report, a composition or a short story. We apply script-level evaluation by concatenating the two answers and using a special $answer\_end$ token to separate the answers in the same script.

The writing errors committed in the scripts are manually annotated using a taxonomy of $80$ error types~\cite{nicholls2003cambridge} together with suggested corrections. An example of error annotations is:
\begin{quote}
\textit{The problems started} $<$e type=``RT"$>\\<$i$>$\textit{in}$<$/i$><$c$>$\textit{at}$<$/c$><$/e$>$ \textit{the box office.}
\end{quote}
where $<$i$><$/i$>$ is the error, $<$c$><$/c$>$ is the suggested correction and the error type \textit{``RT"} refers to \textit{``replace preposition"}.  
For error-oriented models, a word is considered an error if it occurs inside an error tag and the correction is retrieved according to the correction tag.
\begin{table*}[]
\centering
\begin{tabular}{|c|c|c|c|}
\hline
Bootstrapping Model                & Pearson ($r$) & Spearman ($\rho$) & RMSE  \\ \hline
Google Word2Vec 300d & 0.626   & 0.567    & 4.930 \\ \hline
GloVe 50d            & 0.568   & 0.518    & 5.200 \\ \hline
SSWE                 & 0.624   & 0.583    & 4.872 \\ \hline
ESWE                  & 0.667   & 0.637    & \textbf{4.536} \\ \hline
ECSWE                 & \textbf{0.674}   & \textbf{0.642}    & 4.692 \\ \hline
\end{tabular}
\caption{AA results when bootstrapped from different word embeddings and trained on the extended FCE version (FCE$_{\textrm{ext}}$). The bold values indicate the best results.}
\label{table3}
\end{table*}

We train the models on the released public FCE dataset which contains $1,141$ scripts for training and $97$ scripts for testing. In order to examine the effects of training with extra data, we conduct experiments where we augment the public set with additional FCE scripts and refer to this extended version as FCE$_{\textrm{ext}}$, which contains $9,822$ scripts. We report the results of both datasets on the released test set. The public FCE dataset is divided into $1,061$ scripts for training and 80 for development while for FCE$_{\textrm{ext}}$, $8,842$ scripts are used for training and $980$ are held out for development. The only data preprocessing employed is word tokenization which is achieved using the Robust Accurate Statistical Parsing (RASP) system~\cite{briscoe2006second}. 
\vspace{0.2cm} \\ 
\noindent{\bf Training.} 
Hyperparameter tuning is done for each model separately. The SSWE, ESWE and ECSWE models are initialized with GloVe ($d^{wrd} = 50$) vectors, trained for $20$ epochs and the learning rate is set to $0.01$. For SSWE, $\alpha$ is set to $0.1$, batch size to $128$, the number of randomly generated counterparts per ngram to $20$ and the size of hidden layer to $100$.\footnote{Using the same parameters as~\citet{alikaniotis2016automatic}.} For the AA network, initialized with any of the $5$ models, $h$ is set to $100$, and learning rate to $0.001$ when training on public FCE and $0.0001$ on FCE$_{\textrm{ext}}$. The sizes used for error and script filters are shown in Table~\ref{table1}.\footnote{Tuning the filter sizes was done for each model separately; for the Glove and Word2Vec models, a filter of size $3$ performed better than $9$, on both datasets.}
All the networks are optimized using Stochastic Gradient Descent (SGD). 
The AA system is regularized with $L2$ regularization with rate = $0.0001$ and trained for $50$ epochs during which performance is monitored on the dev sets. Finally, the AA model with the best mean square error over the dev sets is selected.

\section{Results and Discussion}
The public FCE results shown in Table~\ref{table2} reveal that AA-specific embedding pre-training offers further gains in performance over the traditional embeddings trained on large corpora (Google and GloVe embeddings), which suggests that they are more suited for the AA task. The table also demonstrates that the ESWE model outperforms the SSWE one on correlation metrics, with a slight difference in the RMSE value. While the variance in the correlations between the two models is noticeable and suggests that the ESWE model is a more powerful one, the RMSE values weaken this assumption. This result could be attributed to the fact that public FCE is a small dataset with sparse error representations and SSWE is trained on $20$ times more data as each ngram is paired with $20$ randomly generated counterparts. Therefore, a more relevant comparison is needed and could be achieved by either training on more data, as will be discussed later, or further enriching the embedding space with corrections (ECSWE). Table~\ref{table2} demonstrates that learning from the errors and their corrections enhances the error pre-training performance on public FCE which indicates the usefulness of the approach and its ability to mitigate the effects of data sparsity. According to the results, training the model based on naturally occurring errors and their correct counterparts is better suited to the AA task rather than introducing artificial noisy contexts and tuning the embeddings according to scripts' scores or relying on word distributions learned from large corpora.  

For a more robust analysis, we examine the performance when training on additional data (FCE$_{\textrm{ext}}$) as shown in Table~\ref{table3}. Comparing the results in Tables ~\ref{table2} and ~\ref{table3} proves that training with more data boosts the predictive power of all the models. It is also clear from Table ~\ref{table3} that with more data, the discrepancy in the performance between SSWE and ESWE models becomes more prominent and ESWE provides a superior performance on all evaluation metrics which suggests that, qualitatively, learning from learners' errors is a more efficient bootstrapping method. However, with FCE$_{\textrm{ext}}$, the ECSWE approach outperforms ESWE on correlation metrics while giving a worse RMSE value. This change in the results when training on a bigger dataset indicates that the effect of incorporating the corrections in training becomes less obvious with enough data as the distribution of correct and incorrect ngrams is enough to learn from. 

\section{Analysis} 
We conduct further analysis to the scores predicted by AA-specific embeddings by investigating the ability of the ESWE and SSWE models to detect errors in text. We run each model for $20$ epochs on the public FCE (ngram size = $3$) and FCE$_{\textrm{ext}}$ (ngram size = $9$) training sets, then test the models on the respective dev sets and examine the output predictions. For simplicity, we assign a binary true score for each ngram with a zero value if it contains any errors and one otherwise. ESWE predicts a score $\in[0,1]$ for each ngram indicating its correctness and hence could be used directly in the evaluation. On the other hand, SSWE predicts two scores for each ngram: \textit{correct score} that it maximizes in comparison to the noisy ngrams and \textit{script score} that should be high for good ngrams that occur in highly-graded scripts. The two scores are hence expected to be high for high-quality ngrams and low otherwise, which suggests that they can be used as proxies for error detection. We calculate the ngram predicted score of the SSWE model as a weighted sum of the correct and script scores, similar to its loss function (Equation \ref{ssweloss} with $\alpha=0.1$), and map the output to a $[0,1]$ probability based on the minimum and maximum generated scores.\footnote{Different score combinations were implemented including using only one score, but they all led to similar results.} We calculate the average precision (AP) between the true scores and predicted ones with respect to the error representing class (true score $=0$) and compare it to a random baseline, where random probability scores are generated. The results are displayed in Table~\ref{table4} which shows that ESWE achieves a higher AP score on all evaluation sets, particularly with public FCE, and SSWE's performance is similar to the random baseline. This result is expected since the ESWE model is trained to predict actual errors, yet an empirical verification was required. We conclude from this analysis that tuning the embeddings based on training writing errors increases their sensitivity to unseen errors which is key for learners' data assessment and yields better performance than comparable pre-training approaches.
\begin{table}[]
\centering
\begin{tabular}{|c|c|c|}
\hline
Model & Public FCE   & FCE$_{\textrm{ext}}$ \\ \hline
Random Baseline & 0.258 & 0.494  \\ \hline
SSWE  & 0.251 & 0.480   \\ \hline
ESWE   & \textbf{0.472} & \textbf{0.539}   \\ \hline
\end{tabular}
\caption{AP results of the random baseline and SSWE and EWE models when trained on public and extended FCE sets and tested on the respective dev sets. The AP is calculated with respect to the error class.}
\label{table4}
\end{table}

\section{Conclusion and Future Work}
In this work, we have presented two error-oriented approaches to train the word embeddings used by writing assessment neural networks. The first approach learns to discriminate between good and bad ngrams by leveraging writing errors occurring in learner data. The second extends the first by combining the error representations with their suggested corrections and tuning the embedding space accordingly. Our motivation for applying these models is to provide neural assessment systems with the minimum features useful for the task in an attempt to boost their performance on challenging datasets while still avoiding heavy feature engineering.
The presented results demonstrate that our error-oriented embeddings are better suited for learners' script assessment than generic embeddings trained on large corpora when both are used to bootstrap a neural assessment model. Additionally, our embeddings have yielded superior performance to those that rely on ranking correct and noisy contexts as well as words' contributions to the script's overall score. Furthermore, extending our error embeddings with error corrections has enhanced the performance when trained on small data, while having a less obvious effect when trained on greater amounts of data which shows their efficacy to enrich the embedding space and mitigate data sparsity issues. We further analysed our embeddings and the score-specific ones and showed empirically that error-oriented representations are better at error detection which explicates their superior performance in learners' data evaluation.

Our best performing model still underperforms the state-of-the-art system by~\citet{yannakoudakis2011new} that utilises a wide variety of features, even when they exclude error related features. However, the improvement obtained by error-oriented models over employing generic embeddings or score-specifc ones suggests that our pre-training approach is a promising avenue of research as it provides neural network assessment with useful information and motivates learning relevant properties associated with language proficiency. 

For future work, it will be interesting to jointly train the score-specific model with the error-oriented one and test if this could further improve the performance. We also suggest fully automating the assessment process by using the outputs of automated error detection and correction systems to build the embeddings rather than relying on handcrafted error annotations. Finally, we encourage further examination for other types of features that could be useful for assessment models and incorporating them in the pre-training stage. This way the performance could be further enhanced with less information than what heavily engineered systems require. 

\bibliography{emnlp2017}
\bibliographystyle{emnlp_natbib}

\end{document}